\documentclass[sigconf,natbib=false]{acmart}
\usepackage{graphicx}
\usepackage{comment}
\usepackage{booktabs}           
\usepackage{amsmath}            
\usepackage{placeins}     
\usepackage{comment}
\usepackage[ruled,vlined]{algorithm2e}

\AtBeginDocument{%
  }

\setcopyright{acmlicensed}
\copyrightyear{2025}
\acmYear{2025}
\acmDOI{XXXXXXX.XXXXXXX}
\acmConference[CODS Dec'25]{Research Track, CODS}{December 17--20,
  2025}{Pune, India}

\acmISBN{978-1-4503-XXXX-X/2018/06}

\RequirePackage[
  datamodel=acmdatamodel,
  style=acmnumeric,
  ]{biblatex}

\begin{document}

\title{Adaptive Financial Sentiment Analysis for NIFTY 50 via Instruction-Tuned LLMs , RAG  and Reinforcement Learning Approaches}

\author{Chaithra}
\affiliation{%
  \institution{National Institute of Technology Karnataka}
  \city{Surathakal}
  \country{India}
}
\email{chaithra.217it001@nitk.edu.in}

\author{Kamesh Kadimisetty}
\affiliation{%
  \institution{Gayatri Vidya Parishad College of Engineering}
  \city{Visakhapatnam}
  \country{India}
}
\email{kameshkadimisetty@gmail.com}

\author{Biju R Mohan}
\affiliation{%
 \institution{National Institute of Technology Karnataka}
 \city{Surathkal}
 \country{India}
}
\email{biju@nitk.edu.in}

\begin{abstract}
Financial sentiment analysis plays a crucial role in informing investment decisions, assessing market risk, and predicting stock price trends. Existing works in financial sentiment analysis have not considered the impact of stock prices or market feedback on sentiment analysis. In this paper, we propose an adaptive framework that integrates large language models (LLMs) with real-world stock market feedback to improve sentiment classification in the context of the Indian stock market. The proposed methodology fine-tunes the LLaMA 3.2 3B model using instruction-based learning on the SentiFin dataset. To enhance sentiment predictions, a retrieval-augmented generation (RAG) pipeline is employed that dynamically selects multi-source contextual information based on the cosine similarity of the sentence embeddings. Furthermore,  a feedback-driven module is introduced that adjusts the reliability of the source by comparing predicted sentiment with actual next-day stock returns, allowing the system to iteratively adapt to market behavior. To generalize this adaptive mechanism across temporal data, a reinforcement learning agent trained using proximal policy optimization (PPO) is incorporated. The PPO agent learns to optimize source weighting policies based on cumulative reward signals from sentiment-return alignment. Experimental results on NIFTY 50 news headlines collected from 2024 to 2025 demonstrate that the proposed system significantly improves classification accuracy, F1-score, and market alignment over baseline models and static retrieval methods. The results validate the potential of combining instruction-tuned LLMs with dynamic feedback and reinforcement learning for robust, market-aware financial sentiment modeling.

\end{abstract}



\maketitle

\section{Introduction}

In financial markets, investor behavior is often influenced by qualitative information, including news headlines, corporate disclosures, and economic prospects \cite{chaithra2024revealing,varghese2022causal}. Extracting sentiment from such textual data provides valuable information for understanding stock movements, guiding portfolio decisions, and developing algorithmic trading strategies. The analysis of sentiment in financial text is a challenging task due to the domain-specific language and the dynamic nature of the market  \cite{chan2017sentiment}.

Large language models (LLMs) have shown good performance in various natural language tasks, but they often fall short in the finance domain \cite{li2023large,inserte2023large,zhang2023instruct}. This is because models typically lack exposure to domain-specific semantics and are not linked to real-world financial feedback; as a result, the sentiment classifications often diverge from actual market responses. News headlines are widely used for sentiment-driven prediction tasks \cite{chen2021stock,bi2021predicting,chen2024deep}. Headlines generally capture the core event and provide stronger, more concentrated sentiment signals than full articles. Although full articles offer detailed background information, they often introduce noise that can dilute predictive performance. However, headlines from certain sources may be extremely brief and lack sufficient contextual information. Therefore, aggregating news from multiple trusted sources is essential to obtain a more comprehensive understanding of the underlying events. The existing approaches have not considered the financial contexts from multiple sources and the reliability of news sources over time.

To address the above issues, we proposed an adaptive sentiment analysis framework that integrates instruction-tuned LLMs with multi-source news data retrieval and a feedback mechanism. We fine-tuned the LLaMA 3.2 3B model using instruction-formatted financial data. We then augmented each input news query with dynamically retrieved context news from trusted financial sources using a retrieval-augmented generation (RAG) approach. The reliability of these sources, determined by source weightage, is adjusted based on the alignment of stock returns.  We also employed a reinforcement learning agent using proximal policy optimization (PPO) to learn weighting strategies that evolve in response to news patterns and stock market behavior.

We created a news dataset comprising NIFTY 50 companies.The sentiment labels are derived from market-adjusted return movements.  Our feedback-aware system outperforms traditional LLMs and static RAG pipelines in both accuracy and alignment with market trends. Through this approach, we tried to bridge the gap between LLM capabilities and financial domain requirements, thereby offering an approach to sentiment analysis that is both context-sensitive and feedback-driven. This approach can be integrated with more powerful LLMs or domain-specific architectures in the future.

The key contributions made by our work can be summarized as follows:
\begin{itemize}
    \item We fine-tuned the LLaMA 3.2 3B model using an instruction-based prompt with domain-specific sentiment data to enable the LLM to understand financial sentiment in an Indian market context.

    \item We constructed a test dataset for NIFTY 50 companies from multiple financial sources, where ground-truth sentiment labels assignment is based on next-day stock returns relative to rolling mean and volatility.

    \item We designed a retrieval-augmented generation (RAG) pipeline that incorporates multi-source contextual evidence, and the top k relevant news articles were retrieved based on cosine similarity of the sentence embeddings.
    
    \item We implemented a direct feedback mechanism that updates the source weights by comparing sentiment predictions with actual next-day stock returns, allowing the system to align more closely with market behavior.
    
    \item We developed a reinforcement learning agent using proximal policy optimization (PPO) to optimize context source weights over time, thereby learning a feedback-aware retrieval strategy that generalizes across days.

\end{itemize}

The remainder of the paper is structured as follows: Section \ref{sec:relatedwork} reviews the literature on financial sentiment analysis and large language models (LLMs). Section \ref{dataset} describes the datasets used in the study.
The section \ref{Methodology} discusses the fine-tuning process, retrieval-augmented generation (RAG), and reinforcement learning. Section \ref{evalution} explains the evaluation metrics and strategy. Section \ref{sec:results} presents the experimental results and analysis. Finally, Section \ref{sec:conclusion} concludes the paper and highlights directions for future research.

\section{Literature Review}
\label{sec:relatedwork}

Financial sentiment analysis has been performed using dictionary-based evaluations, machine learning (ML), deep learning (DL), transformer models like FinBERT, and, more recently, large language models (LLMs). This section surveys key contributions from each phase.

\subsection{Dictionary-Based Approaches}

The earliest sentiment analysis was performed using manually curated sentiment lexicons. Tools like SentiWordNet, VADER, and the Loughran-McDonald Financial Sentiment Dictionary ~\cite{loughran2011liability} provided word-level polarity scores, enabling straightforward sentiment classification by counting positive and negative tokens. While these methods were interpretable and domain-specific to some extent, they lacked contextual understanding and failed in the presence of sarcasm, negations, or complex sentence structures. Their inability to adapt to new language patterns and lack of feedback integration limited their effectiveness in financial domains.

\subsection{Machine Learning Techniques}
Dictionary-based methods often do not consider the context and ignore domain-specific meanings \cite{malo2014good}; therefore, to improve generalization, classical ML models that are trained with sentence label pairs, such as Naive Bayes, support vector machines (SVM), and logistic regression, were employed ~\cite{pang2002thumbs}. These models relied on handcrafted features, such as n-grams, POS tags, and TF-IDF vectors, and outperformed lexicon-based techniques in many scenarios. However, they still struggled with long-range dependencies and syntactic structure.  Moreover, these models were typically static post-training and did not incorporate time-evolving market behavior. 

\subsection{Deep Learning Techniques}
The deep learning models enabled the automatic extraction and modeling of complex linguistic structures \cite{sohangir2018big}. Models such as CNNs, RNNs, and LSTMs \cite{murthy2020text} have been widely used for sentiment classification, as they can capture sequential dependencies in text, making them well-suited for analyzing financial reports and news articles. But these models require more data for training, are domain-agnostic, and tend to overfit, especially in low-resource financial domains. Additionally, these approaches also did not leverage external feedback or context beyond the input sequence.

\subsection{Transformer Models and FinBERT}
The transformer architecture, such as BERT~\cite{devlin2018bert}, with self-attention mechanisms and deep contextual embeddings, has demonstrated improved performance in a wide variety of NLP tasks, including sentiment analysis. FinBERT~\cite{araci2019finbert}, a financial-domain adaptation of BERT, was pre-trained on financial documents, such as SEC filings, and has outperformed deep learning models in finance-specific tasks.  These models are text-only models and lack mechanisms to adapt based on real-world feedback, such as stock market responses.

\subsection{Large Language Models (LLMs)}

 LLMs like GPT-3~\cite{brown2020language}, PaLM, and LLaMA~\cite{touvron2023llama} have shown improvement in zero-shot and few-shot sentiment analysis tasks, but often underperform in finance due to domain-specific text and context \cite{zhang2023instruct}. Instruction-tuned versions of these models have also demonstrated strong performance in sentiment classification \cite{10.1145/3706119}.  In many earlier works, real-time market signals were not incorporated, making them less suitable for adaptive financial decision-making tasks.

\subsection{Instruction Alignment and Feedback-Aware Models}

To bridge this gap, recent research has focused on aligning LLMs not only with human instructions but also with real-world feedback.
Zhang et al \cite{zhang2023enhancing} found that the concise nature of
financial news often lacks sufficient context, which can significantly
decline the reliability of LLM in financial sentiment analysis. To address it, they introduced a retrieval-augmented LLM framework for financial sentiment analysis.
Zhao et al.~\cite{zhao2024aligning} introduced a novel adaptive sentiment framework using Retrieval-Augmented Generation (RAG) pipelines, where external context from multiple sources is combined with LLMs for inference. They incorporated feedback from stock market returns to dynamically adjust source weights via reinforcement learning, thereby making the model context-aware and market-responsive. Our work builds on this paradigm by applying the methodology to the Indian equity market and NIFTY 50 data. We extend it by combining domain-specific fine-tuning of LLaMA 3.2, retrieval with dynamic feedback, and PPO-based reinforcement learning for robust and generalizable source optimization.

\section{Datasets}\label{dataset}

We utilized two primary datasets in different stages of the pipeline:

     \subsection{Fine-Tuning Dataset}
     We utilized \textit{SentiFin} dataset \cite{sentfin2024github} as the primary source for instruction-based fine-tuning of the LLaMA 3.2 3B model. This dataset contains news data for the Indian stock market. The dataset comprises 10,572 labeled headlines across three sentiment classes: \texttt{positive}, \texttt{neutral}, and \texttt{negative}. Each instance includes a short financial headline, the sentiment label, and additional metadata such as the news source and publication date.  The samples are formatted into instruction-based prompt-response pairs to train the LLM. Table~\ref{tab:sentfin-stats} presents the distribution of sentiment labels:

\begin{table}[h]
\centering
\caption{SentiFin Dataset Statistics}
\label{tab:sentfin-stats}
\begin{tabular}{lcc}
\toprule
\textbf{Sentiment Class} & \textbf{Count} &  \\
\midrule
Positive & 4,505 \\
Neutral  & 3,695 \\
Negative & 3,386 \\
\bottomrule
\end{tabular}
\end{table}

We split the dataset into training and testing sets using an 80:20 ratio, ensuring a stratified distribution of sentiment classes.

  \subsection{ Retrieval Augmented Generation (RAG) Dataset and Test Dataset:}
  We created a dataset of ~8,000 news headlines related to NIFTY 50 companies, spanning 2024–2025, to support the RAG pipeline.  This news data is collected through webscraping. All news data was collected from trusted Indian financial sources, such as Yahoo Finance and MoneyControl. The major news source details are given in the Table \ref{tab:RAG}.  Headlines were cleaned using basic NLP preprocessing, which included the removal of HTML tags, normalization of dates and entities, and mapping of tickers to standard NSE symbols.

  \begin{table}[]
      \centering
       \caption{RAG Dataset Statistics}
       \label{tab:RAG}
      \begin{tabular}{cc}
      \toprule
         \textbf{Source}  &\textbf{Count}  \\
         \midrule
          Business Standard &1031\\
        NDTV Profit & 1002\\
        Financial Express & 830\\
        The Economic Times & 743\\
        Mint & 699\\
        MoneyControl & 585\\
        Business Today & 449\\
        ET Now & 335  \\
        \bottomrule
      \end{tabular}

  \end{table}
  
 We also constructed a separate test dataset and generated its ground-truth sentiment labels. The sentiment labeling was carried out using the following strategy.
  
  \begin{itemize}
  \item Daily returns were computed as the percentage change in opening price.
\item For each stock, we computed a 30-day rolling mean and standard deviation of returns.
\item Each headline was aligned with the next trading day’s return. If historical price statistics were unavailable, the label was marked as \textit{unknown}.
\item Sentiment labels were assigned based on the deviation of the next-day return from the rolling statistics:
\begin{itemize}
\item \textbf{Positive:} Return > Mean + Std
\item \textbf{Negative:} Return < Mean – Std
\item \textbf{Neutral:} Otherwise
  \end{itemize}

  \end{itemize}
This objective labeling approach avoids subjective human annotation and grounds sentiment in actual market behavior. The test dataset statistics are given in the Table \ref{tab:test_dataset}.
\begin{table}[h]
    \centering
   
    \caption{Test Dataset Statistics}
     \label{tab:test_dataset}
    \begin{tabular}{lcc}
\toprule
\textbf{Sentiment Class} & \textbf{Count} & \\
\midrule
   Positive & 823 \\
   Negative & 1188 \\
   Neutral  & 4123 \\
   \bottomrule
\end{tabular}
    
    \label{tab:placeholder}
\end{table}

\section{Methodology} \label{Methodology}
The proposed adaptive financial sentiment analysis system consists of the following modular components:

\begin{enumerate}
    \item \textbf{Fine-Tune LLaMA :}   This module instruction-tune LLaMa 3.2 on financial sentiment data to classify news headlines into \textit{positive}, \textit{negative}, or \textit{neutral} sentiment.

    \item \textbf{Retrieval-Augmented Generation (RAG) Module:} Dynamically retrieves top-$k$ context snippets from a corpus of NIFTY 50 financial news using sentence embeddings and cosine similarity, with source-specific weighting.

    \item \textbf{Market Feedback Engine:} Compares model predictions with next-day stock price movements to reinforce or penalize the reliability scores of sources, simulating an environment-aware feedback loop.

    \item \textbf{Source Credibility Weighting Mechanism:}  This module assigns credibility weights to news sources using both rule-based and reinforcement-learning-based optimization strategies.
\end{enumerate}

\begin{figure*}[h]
\centering
\includegraphics[width=\linewidth]{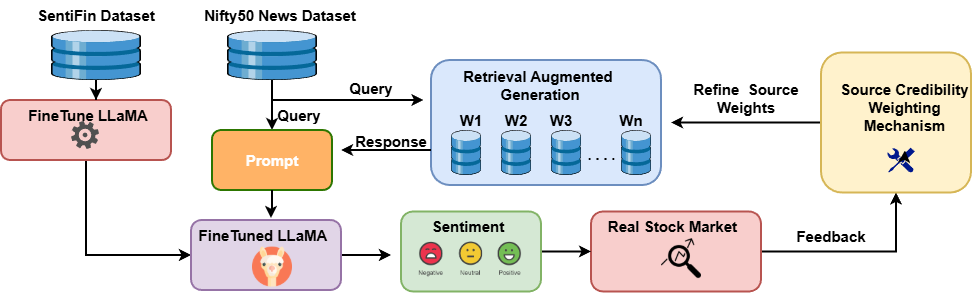}
\caption{Overview of the proposed adaptive sentiment analysis pipeline}
\label{fig:architecture}
\end{figure*}

A visual overview is shown in Figure~\ref{fig:architecture}.

\subsection{Instruction-Based Fine-Tuning of LLaMA 3.2}

We fine-tuned the \texttt{unsloth/Llama-3.2-3B-Instruct \footnote{https://huggingface.co/unsloth/Llama-3.2-3B-Instruct}} model using the quantized low-rank adaptation (QLoRA) technique, which enables memory-efficient training on GPUs.  The finetuning process is outlined below.

\begin{itemize}
    \item \textbf{Instruction Formatting:} The training samples are formatted in the  LLaMA Instruct format. The structure is:

\begin{verbatim}
<s>[INST] Classify the sentiment of the following 
financial sentence: [headline] [/INST] [label]</s>
\end{verbatim}

    \item \textbf{Model and Adapter Configuration:}  The Unsloth implementation is used for memory-optimized fine-tuning. LoRA adapters are applied to the attention and MLP layers (\texttt{q\_proj}, \texttt{k\_proj}, \texttt{v\_proj}, \texttt{o\_proj}, \texttt{gate\_proj}, \texttt{up\_proj}, \texttt{down\_proj}), with a rank of 16 and LoRA alpha set to 16.

    \item \textbf{Training Details:} HuggingFace's \texttt{SFTTrainer} with 4-bit quantization, mixed precision (fp16), a batch size of 4 (with gradient accumulation), and a learning rate of 2e-5 is used for training. The model is trained for 3 epochs and saves the best-performing checkpoints to disk.

    \item \textbf{Evaluation:} The model is validated using accuracy and F1-score on a held-out test set. 

\end{itemize}

This instruction-based fine-tuning enhances the model’s ability to perform financial sentiment classification in the Indian stock market.

\subsection{Retrieval-Augmented Generation (RAG)}
To provide additional context for the headline with relevant and sentiment-aligned background information, we implemented a Retrieval-Augmented Generation (RAG). It consists of the following key components:
\begin{enumerate}
    \item \textbf{Candidate Document Retrieval:} Given a test headline, stock symbol, and date, retrieve a set of candidate news articles from a RAG dataset, within a 3-day window centered around the test date. It retrieves from sources with non-zero reliability weights.

    \item \textbf{Sentiment-Aware Filtering:} Filter retrieved documents based on sentiment cues present in the headline, such as "gain", "fall", "stable". This process ensures contextual alignment and minimizes noise.

    \item \textbf{Sentence Embedding and Scoring:} The headline and candidate headings are embedded using the \texttt{all-MiniLM-L6-v2} model. Cosine similarity scores are computed between the headline and each candidate. These are then multiplied by the corresponding source reliability weights to get a weighted relevance score.

    \item \textbf{Top-k Context Selection and Prompt Construction:} The top-ranked documents are selected based on the weighted relevance score. These are concatenated with the original headline to form a single instruction-formatted prompt, which is then passed to the fine-tuned LLaMA model for final sentiment prediction.

\end{enumerate}

This adaptive RAG layer enables the model to retrieve external context selectively from reliable sources, thereby improving classification accuracy on ambiguous or context-poor headlines.

\subsection{ Source Credibility Weighting Mechanism} 
Credibility weights for news sources are assigned using dynamic source weighting, based on market feedback and reinforcement learning, as described in the following sections.
\subsubsection{Dynamic Source Weighting via Market Feedback}

To validate the credibility of various financial sources, we employed a dynamic source weighting approach guided by real-world market feedback. Initially, each context source, such as the Economic Times and Moneycontrol, is assigned a normalized reliability weight. During inference, retrieved documents from these sources contribute additional context to the prompt, and the final sentiment prediction is compared with the true market response.

After each prediction:

\begin{itemize}
    \item If the predicted sentiment aligns with the actual stock movement, for example, a positive sentiment followed by a stock price increase, the weights of the contributing sources are increased.
    \item Conversely, misaligned predictions lead to penalization of the associated sources by reducing their weights.
    \item A neutral return zone ±0.5\% is defined to avoid penalizing predictions where the price change is not significant.
\end{itemize}

Weight updates are performed by a lightweight gradient-style rule:
\begin{equation}
w_{\text{new}} = \text{clamp}(w_{\text{old}} \pm \alpha)
\end{equation}
where $\alpha$ is a small learning rate of $1 \times 10^{-4}$, and weights are normalized after each update to ensure they sum to 1. This feedback loop iterates over the entire evaluation dataset. It supports sources that align with observed stock trends and demotes those that frequently contribute to incorrect predictions.

By integrating this mechanism into our RAG pipeline, the model continuously adjusts its source reliability, thereby learns a context selection strategy that is sensitive to historical prediction performance and market behavior.

\subsubsection{Dynamic Source Weighting via Reinforcement Learning}
The direct market feedback updates source weights, but it leads to short-term, unstable adjustments. To enable more robust generalization across unseen data and source combinations, we model the source weighting process as a reinforcement learning (RL) task and apply proximal policy optimization (PPO). The PPO's surrogate clipped objective provides stable and conservative policy updates, preventing divergence in dynamic or noisy environments \cite{siboo2023empirical}. The key components of PPO-based reinforcement learning are:

\begin{itemize}
    \item \textbf{Environment:} In a reinforcement learning environment, each episode corresponds to a sequence of sentiment prediction tasks. At each step, the environment provides a financial headline, its symbol, and date, and the agent selects a weight distribution over the context sources.
    
    \item \textbf{State:} The state of the PPO agent consists of the current normalized source weights, combined with features from the query headline or past performance.
    
    \item \textbf{Action:} The agent outputs a normalized vector representing updated weights for context sources within the RAG pipeline.
    
    \item \textbf{Reward:} In the reward function, a reward of +1 is assigned if the predicted sentiment aligns with the ground-truth label derived from next-day stock returns, and -1 otherwise. A neutral return zone can be used to clip insignificant reward effects.
\end{itemize}

The PPO agent is trained across multiple time steps to maximize the cumulative reward. Agent learns an adaptive strategy for weighting sources based on both the current context and long-term trends. The PPO agent training loop interacts with the environment by performing document retrieval, constructing prompts, and inferring models at each step, thereby it simulates a full pipeline execution. After training, the agent's final policy is used to generate the optimized source weight vector. This weight vector is then applied to evaluate the full test set. The advantage of this approach is that it enables exploration across sources and stabilizes the retrieval mechanism under uncertain market conditions. The details of weight refinements using PPO  is given in the Algorithm \ref{alg:ppo}.

\begin{algorithm}
\caption{ RAG Source Weights using PPO}
\label{alg:ppo}
\KwIn{Initial weights $\mathbf{w}_0$, Test set $\mathcal{D}_{test}$, PPO hyperparameters $(\eta, \epsilon)$}
\KwOut{Optimized source weights $\mathbf{w}^*$}
\BlankLine
Initialize policy $\pi_\theta$ and value $V_\phi$\;
Initialize environment state $s_0 = (\mathbf{w}_0, \text{accuracy history}, \text{market indicators})$\;
\While{not converged}{
  Sample batch of trajectories:
  \begin{enumerate}
    \item Select action $\mathbf{a}_t = \pi_\theta(s_t)$ representing new source weights
    \item Normalize $\mathbf{a}_t$ so $\sum_i a_{t,i}=1$
    \item Perform RAG inference with $\mathbf{a}_t$ to predict $\hat{y}_t$
    \item Compute reward $r_t = +1$ if $\hat{y}_t = y_t$, else $-1$
    \item Observe next state $s_{t+1}$
  \end{enumerate}
  Estimate advantage $A_t = r_t + \gamma V_\phi(s_{t+1}) - V_\phi(s_t)$\;
  Update policy by maximizing PPO clipped objective:
  \[
  L^{\text{CLIP}}(\theta) = \mathbb{E}_t\!\left[\min\left(r_t(\theta)A_t,\ \text{clip}(r_t(\theta),1-\epsilon,1+\epsilon)A_t\right)\right]
  \]
  Update value network by minimizing mean squared error loss\;
}
Return $\mathbf{w}^* = \mathbb{E}[\pi_\theta(s)]$\;
\end{algorithm}

\section{Evaluation and Metrics} \label{evalution}
This section discusses the evaluation pipeline and the metrics used for the evaluation.

\subsection{Evaluation Pipeline}
We evaluated our system using both the base instruction-tuned model and the full RAG-enhanced + PPO-optimized pipeline. The evaluation was conducted using a labeled test set.

\begin{itemize}
\item For each headline in the test set, the top-$k$ relevant context passages were retrieved via the RAG module using source-weighted cosine similarity.
\item A sentiment classification prompt was constructed using the headline and selected context.
\item The fine-tuned LLaMA 3.2 model generated the predicted sentiment.
\item Predictions were made with both market feedback–adjusted weights and PPO-optimized weights.
\end{itemize}

\subsection{Metrics}

We used the following standard classification metrics:

\begin{itemize}
\item \textbf{Accuracy:} The percentage of correctly predicted labels.
\begin{equation}
    Accuracy=\frac{TP+TN}{TP+TN+FP+FN}
\end{equation}
\item \textbf{F1-Score:} The harmonic mean of precision and recall, averaged across \texttt{positive}, \texttt{neutral}, and \texttt{negative} classes.
Let
\begin{itemize}
    \item $K$ = number of classes
    \item $n_c$ = support of class $c$
    \item $N = \sum_{c=1}^{K} n_c$
    \item $F1_c$ = F1-score for class $c$
\end{itemize}
\begin{equation}
Weighted\ F1= \sum_{c=1}^{K} \frac{n_c}{N} \cdot F1_c 
\end{equation}
\[
F1_c = \frac{2 \cdot P_c \cdot R_c}{P_c + R_c}
\]

\[
P_c = \frac{TP_c}{TP_c + FP_c}, \qquad
R_c = \frac{TP_c}{TP_c + FN_c}
\]
\end{itemize}

We also compare performance between:

\begin{enumerate}
\item Instruction-tuned model (no context).
\item RAG-enhanced predictions with static weights.
\item RAG with market feedback–updated weights.
\item RAG with PPO-learned dynamic weighting.
\end{enumerate}
 We also incorporated short-term stock price movements by using price information from the preceding three trading days in addition to the news data. The price information is converted into a natural language form. The generated price descriptions were then appended to the multisource news context for predictions.


\section{Results}
\label{sec:results}

This section presents the experimental results obtained on a custom-labeled dataset of NIFTY 50 financial news headlines. The evaluation encompasses various stages of the sentiment analysis pipeline, including instruction-tuned LLM, retrieval-augmented generation (RAG) with cosine similarity, market feedback-based source reweighting, and PPO-optimized weighting. Table~\ref{tab:results_table} summarizes the performance across all configurations using accuracy and weighted F1-score metrics. 

\begin{table*}[h]
\centering
\caption{Performance comparison across model variants}
\label{tab:results_table}
\begin{tabular}{lcc}
\toprule
\textbf{Model Variant} & \textbf{Accuracy} & \textbf{Weighted F1} \\
\midrule

Fine-Tuned LLaMA 3.2 & 0.5520 & 0.5375 \\
RAG + Without Market Feedback &   0.6094 & 0.5722 \\
RAG + With Market Feedback (Cosine Similarity) & 0.5999 & 0.5705 \\
RAG + With Market Feedback (WOC) & 0.6153 & 0.5746\\
RAG + PPO Optimized Weights &  0.6109 & 0.5733 \\

\midrule
\textbf{Price Context}& \\
RAG + Without Market Feedback   &0.6619 &  0.5677\\
RAG + With Market Feedback (Cosine Similarity) & 0.6612 & 0.5668  \\
RAG + With Market Feedback (WOC) &0.6650  & 0.5674 \\
RAG + PPO Optimized Weights & 0.6630 &  0.5683 \\

\midrule
\textbf{Baseline Methods}& \\
FinBERT \cite{araci2019finbert} & 0.4852 & 0.5027 \\   
RoBERTa &   0.5800 & 0.5551   \\
\bottomrule
\end{tabular}
\end{table*}

\paragraph{Fine-Tuned LLaMA:} The instruction-tuned LLaMA model achieves an accuracy of 0.5520  and a weighted F1-score of 0.5375, with the highest recall in the \texttt{neutral} class. The results indicate that the model struggles to accurately identify \texttt{positive} and \texttt{negative} sentiments, reflecting the ambiguity in financial headlines without additional context.

\paragraph{RAG + Without Market Feedback:} The RAG-based contextual retrieval using cosine similarity scores resulted in an accuracy of 0.6094 and an F1-score of 0.5722. This shows the importance of augmenting inputs with relevant financial context.

\paragraph{RAG + With Market Feedback:} The weighted overlap coefficient (WOC) approach \cite{zhao2024aligning} has achieved better performance, with an accuracy of 0.6153 and an F1 score of 0.5746. If the query size is small, the WOC approach often outperforms the cosine similarity-based approach. The penalizing or rewarding of each source's contributions based on sentiment-return match improves the robustness of context aggregation.

\paragraph{PPO-Optimized Weights:}The PPO agent learns optimal source weight distributions with similar accuracy and F1-score to the feedback model, but with a more refined source selection strategy, as shown in Figure~\ref{fig:ppo_weights}. While the improvement in aggregate metrics is marginal, PPO generally helps generalize retrieval preferences over unseen stocks and time periods.

\paragraph{Price Context :} With the addition of price context, all approaches achieved an accuracy of approximately 0.66. However, the weighted F1 score did not improve. The higher accuracy is largely driven by the model’s increased prediction of the majority neutral class. Therefore, the price data, as a context, did not help in predicting the minority class.

\paragraph{FinBERT and RoBERTa :} The FinBERT model, although pretrained on financial text, was unable to predict the class labels correctly, whereas the RoBERTa model achieved better performance compared to FinBERT. This suggests that incorporating additional context from multiple news sources and prices enhances model performance.

\subsection{Source Weight Analysis}
\begin{figure}[h]
\centering
\includegraphics[width=0.95\linewidth]{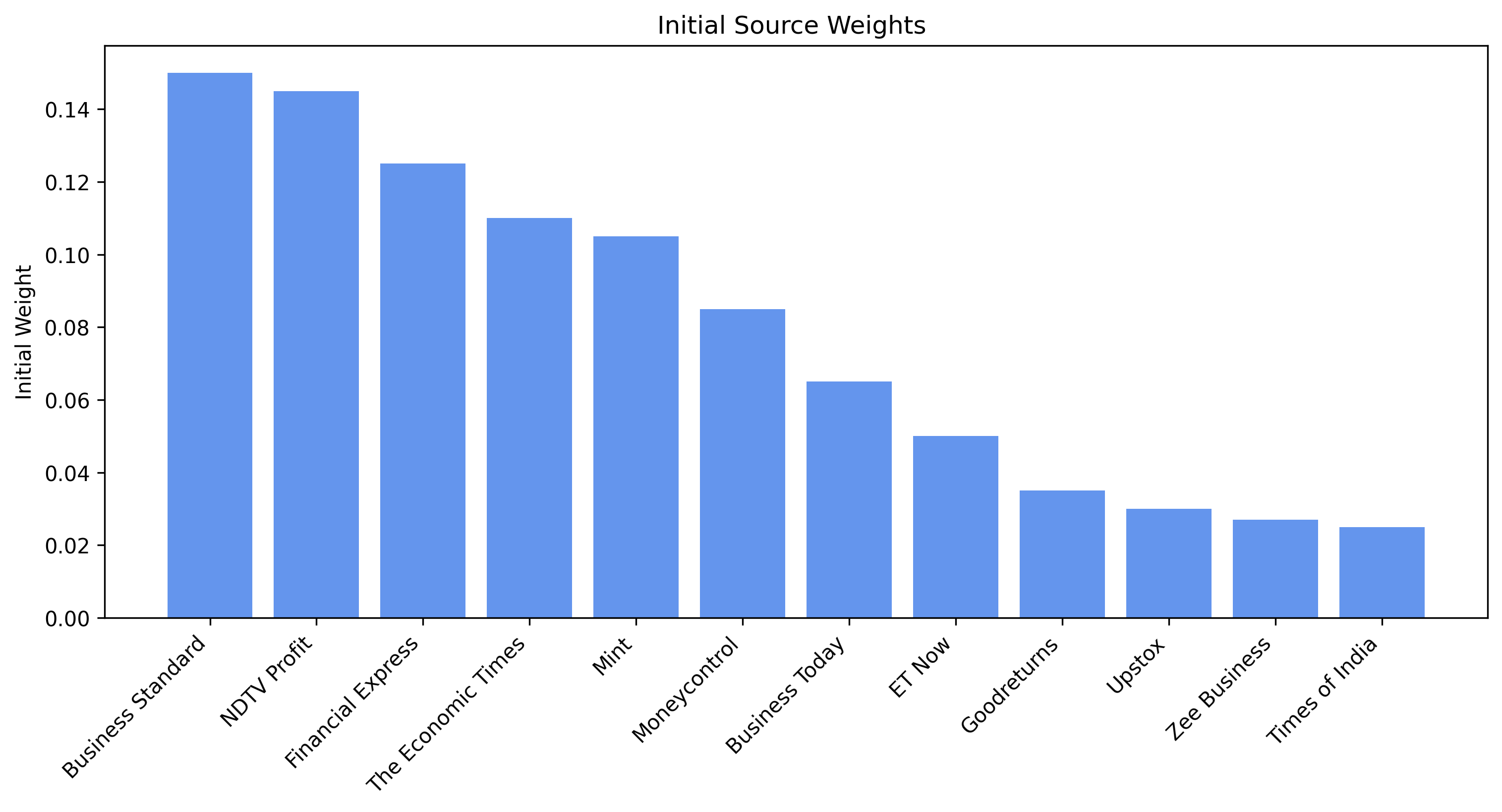}
\caption{Initial normalized source weights (manually initialized)}
\label{fig:initial_weights}
\end{figure}

\begin{figure}[h]
\centering
\includegraphics[width=\linewidth]{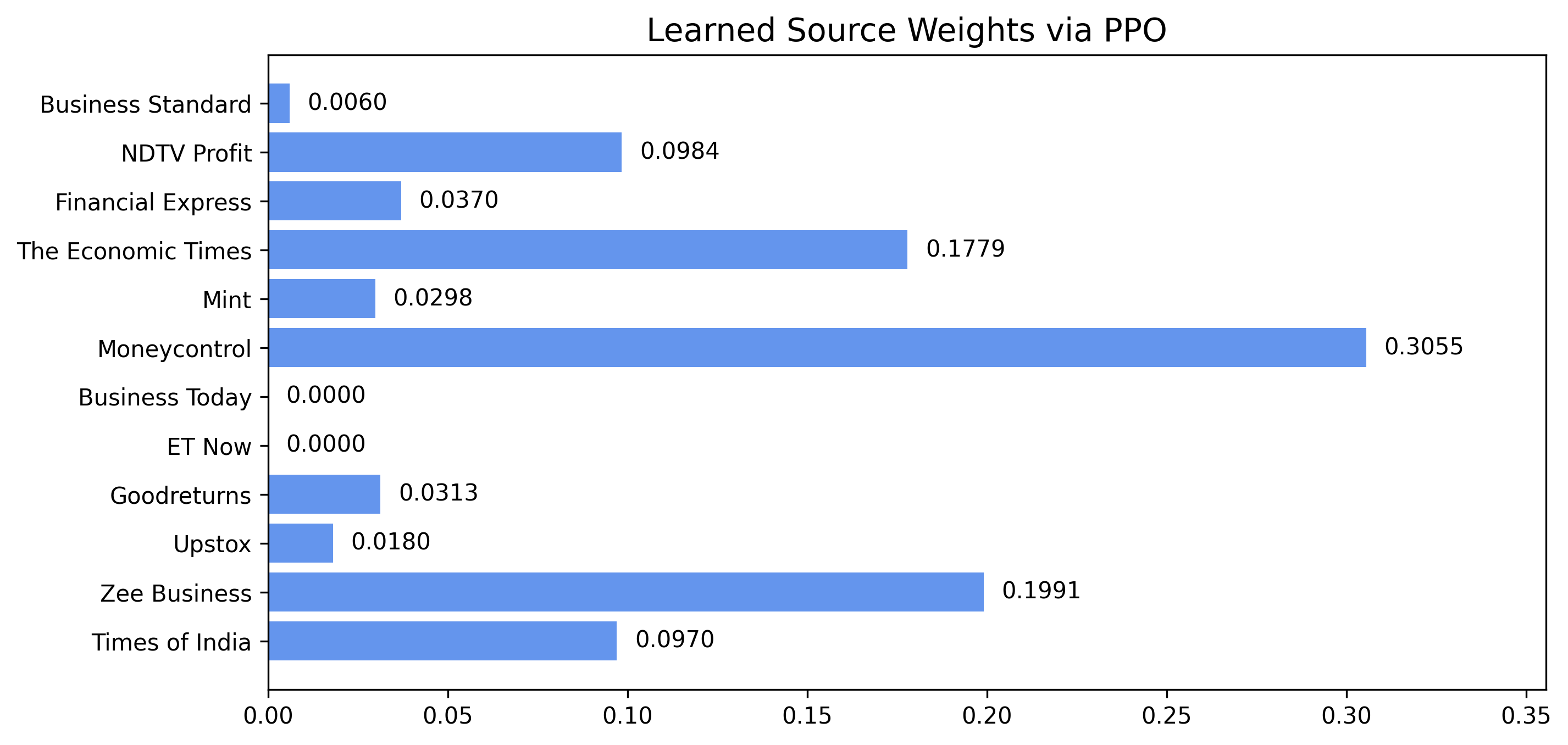}
\caption{Final source weights learned via PPO optimization}
\label{fig:ppo_weights}
\end{figure}

The distribution of source weights before and after PPO optimization is illustrated in Figure \ref{fig:initial_weights} and \ref{fig:ppo_weights} to understand how the system adapts its information sourcing strategy.

Initially, all sources were assigned weights based on perceived editorial reliability, popularity, and market coverage. For instance, Business Standard and NDTV Profit were assigned the highest initial importance of 15.23\% and 14.80\%, respectively, followed by Financial Express, The Economic Times, and Mint. These manual assignments reflected conventional assumptions about the credibility of financial news sources.

After training the PPO agent over sentiment-return alignment feedback, a significant redistribution of weights was observed. The agent learned to suppress sources that contributed less reliably to sentiment classifications. Moneycontrol, Zee Business, and The Economic Times emerged as the most influential sources, while previously dominant sources such as NDTV Profit and Business Standard were assigned a weight of zero.

This weight reallocation indicates a trust profile for each source based on empirical usefulness in sentiment prediction. The learned weights implicitly capture patterns such as news article writing style, topic specification details, and reward sources whose contextual data improved downstream sentiment-return accuracy.

\section{Conclusion}
\label{sec:conclusion}

In this work, we proposed an adaptive framework for financial sentiment analysis by integrating large language models and real-world stock market feedback. We finetuned the LLaMA 3.2 3B model using the SentiFin dataset to adapt to the financial domain. To incorporate context from a diverse set of financial news sources, the RAG pipeline was proposed. The RAG pipeline has enhanced prediction accuracy compared to baseline models.
In addition to supplying contextual information, source trustworthiness was incorporated through the market feedback mechanism. Incorporating these dynamically weighted sources improved prediction performance compared to using static initial weights. To further generalize and optimize source weights, the reinforcement learning algorithm Proximal Policy Optimization (PPO) is used. The performance gain was marginal, but it led to a refined source selection process.

The experiments conducted on NIFTY 50 news data demonstrate that each component, from instruction tuning to retrieval through RAG to reward-based adaptation, makes a significant contribution to the system’s overall performance. The proposed model not only achieved prediction accuracy but also robustness to news variation and stronger alignment with stock market behavior. The current limitation of our approach is its dependence on multi-source news data. Although price information from the preceding days was incorporated, it improved accuracy but did not improve the F1 scores. In many cases, news sentiment alone may not capture stock behavior, especially when price movements are driven by stock fundamentals. Therefore, incorporating stock fundamental indicators and peer stocks from the same industry may help improve the alignment of sentiment and the actual market.

\printbibliography

\end{document}